\documentclass[letterpaper, 10pt, conference]{IEEEtran}
\usepackage{amsmath,amsfonts}
\usepackage{amsthm}
\usepackage{amssymb}
\usepackage{xpatch}
\usepackage{algorithm}
\usepackage{algpseudocode}
\usepackage{array}
\usepackage[caption=false,font=normalsize,labelfont=sf,textfont=sf]{subfig}
\usepackage{textcomp}
\usepackage{stfloats}
\usepackage{url}
\usepackage{verbatim}
\usepackage{graphicx}
\usepackage{tabularx}
\usepackage{cite}
\usepackage{url}
\usepackage[
top    = 0.694in,
bottom = 1.05in,
left   = 0.64in,
right  = 0.64in]{geometry}

\IEEEoverridecommandlockouts
\xpatchcmd{\proof}{\hskip\labelsep}{\hskip3\labelsep}{}{}

\makeatletter
\xpatchcmd{\proof}{\@addpunct{.}}{\@addpunct{:}}{}{}
\makeatother

\begin{document}

\title{\LARGE \bf
	Behavior Tree-Based Task Planning for Multiple Mobile Robots using a Data Distribution Service}

\author{Seungwoo Jeong,
	Taekwon Ga,
	Inhwan Jeong,
	Jongeun Choi,~\IEEEmembership{Member,~IEEE}

\thanks{This study was supported by Hyundai Robotics. All robots for testing, demonstration, and environmental settings were operated in the Hyundai Robotics Engineering laboratory.

Seungwoo Jeong, Taekwon Ga, and Jongeun Choi are with the School of Mechanical Engineering, Yonsei University, Seoul 03722, Republic of Korea (e-mail: slsw@yonsei.ac.kr; taek111@yonsei.ac.kr; jongeunchoi@yonsei.ac.kr). Inhwan Jeong is with Hyundai Robotics, Yong-in 16891, Republic of Korea (e-mail: 123inani@hyundai-robotics.com). The corresponding author is Jongeun Choi.}
}

\maketitle

\begin{abstract}
In this study, we propose task planning framework  for multiple robots that builds on a behavior tree (BT). BTs communicate with a data distribution service (DDS) to send and receive data. Since the standard BT derived from one root node with a single tick is unsuitable for multiple robots, a novel type of BT action and improved nodes are proposed to control multiple robots through a DDS asynchronously. To plan tasks for robots efficiently, a single task planning unit is implemented with the proposed task types. The task planning unit assigns tasks to each robot simultaneously through a single coalesced BT. If any robot falls into a fault while performing its assigned task, another BT embedded in the robot is executed; the robot enters the recovery mode in order to overcome the fault. To perform this function, the action in the BT corresponding to the task is defined as a variable, which is shared with the DDS so that any action can be exchanged between the task planning unit and robots. To show the feasibility   of our framework in a real-world application, three mobile robots were experimentally coordinated for them to  travel alternately to four goal positions  by the proposed single task planning unit via a DDS.
\end{abstract}


\section{Introduction}
Recently, task planning for multiple mobile robots has been actively investigated. Task planning is a type of programming for tasks assigned to robots, which is typically performed using a robot programming language \cite{8793875}. 
A robot programming language has a language parser that interprets a text-based language model to implement a machine language that is suitable for commands to robots. With the aid of diagram-based programming language research, task planning based on a finite state machine (FSM) has emerged as a popular methodology \cite{9519714, 9180351}. A block programming language supported by a graphical user interface has also been introduced and applied to cooperative robots \cite{9207834}. A robot language based on sequential texts is suitable for an interpreted robot machine language. However, such a language is usually dependent on robot hardware, which can be executed only within a robot. Since single task planning is only applicable to one robot, programming for multiple robots is necessary for reducing the effort required to perform individual task planning using a separate programming language for each robot.

Let us consider a scenario in which  a supervisory control unit controls multiple robots that cooperate with each other. A single control unit for controlling multiple robots that can issue commands while observing the states of all robots is highly desirable. It is clear that it would be very efficient to operate a single control unit, rather than using multiple control units for issuing commands to each robot. In previous studies \cite{1372526, 4097991, 891049, 6008651, 1470318, xu2016bayesian, choi2009distributed}, a control unit was established to control multiple robots through a communication network. Typically, different network addresses for $N$ robots are assigned to identify individual robots which are managed using TCP/IP or a sensor network. However, such a control unit must know the network address of each robot. In this case, if a robot is powered down or if another robot enters a multi-robot cluster for replacement, the control unit must respond to new robots individually. Therefore, in this paper, we propose the DDS connection method for multiple robots to replace  peer-to-peer connection, which has been studied extensively in the past. Moreover, we propose a novel BT framework using a DDS that is easy to manage on a single control unit.

The BT was originally designed to replace the FSM in the game industry \cite{8319483, 8116624, 7435292, 5325892}. An FSM created to model discrete event systems is essentially a cyclic graph data structure that represents the state to consider for a specific event occurring in an external environment. An FSM provides an intuitive interface, but it has significant shortcomings when attempting to model complex states and actions. Furthermore, when adding new states and actions to an FSM, the complexity increases exponentially according to the topological complexity of the existing FSM \cite{BTinRoboticsAI}. Many single-robot or multi-robot control schemes have been studied based on FSMs and many studies on modular have been performed, but reactive robot control for dynamic environments has recently emerged based on BTs. Additionally, it has been demonstrated that a BT is more suitable for real-time operations than other methods \cite{7790863}. A BT can model sequential behavior compositions, perceptiveness, safe modularity, and decision trees for hybrid control \cite{7790863}. Furthermore, some works have demonstrated that linear temporal logic \cite{8977342} and control barrier functions \cite{9304151} can also be implemented using a BT.

Since BTs have been applied to control various engineering and robotic applications \cite{sprague2018improving, 8594319, 8594083, 8206598, styrud2021combining}, many new types of BTs have been developed. A novel approach to providing reactivity in dynamic environments was introduced with the back-chaining concept \cite{9145620}, where an action incorporates both preconditions and postconditions \cite{8794128}. If the preconditions are not met, then the most appropriate action is selected to achieve the goal repeatedly. Some studies have added memory functions to BTs. For example, a flight control computer was constructed by introducing memory variables into a BT \cite{8967928}. Memory variables were maintained by a latch that remembered the successes or failures returned by subtrees and did not reevaluate extra subtrees without a reset. Another memory-type BT implemented an algorithm to skip unchanged subtrees by placing the indexes of child trees separately in some control flow nodes \cite{9448466}. In a similar approach, a skipper node was placed in a parent node to determine whether to execute a child node \cite{9341562}. In addition to BTs with memory, another type of research on concurrent BTs attempts to solve the issues associated with the notions of progress and resource sharing, where non-independent BTs that influence each other are executed in parallel \cite{8593504}. Unlike BT for single robots, there are BT studies on multiple robots. The study for operating swarm robots to implement cooperative strategies applied BT to manage multiple kilobots \cite{jones2018evolving}. 
Another similar study applied multiple BTs that separates the local task from the global task by adding a fault-tolerant function \cite{7558452}.

In contrast to  studies \cite{BTinRoboticsAI, sprague2018improving, 8594319, 8206598, styrud2021combining, 8794128, 8967928} that planned multiple tasks for only one robot using a single BT, we extend the BTs to the multiple robots. The main contributions of this paper are summarized as follows. 
\begin{itemize}
\item {We show how to use  multiple BTs to operate multiple tasks of multiple robots in a separate task planning unit. The separate planning unit is regarded as supervisory task planning unit.} 
\item Additionally, The supervisory task planning via a DDS domain are experimentally validated with three mobile robots. In contrast to \cite{jones2018evolving}, the supervisory task planning unit communicates multiple robots via DDS rather than the individual BT to control a single robot communicates each other. 
\item The task planning and fault tolerant function were completely separated. Each function is implemented with the BT. Compared to the study from \cite{7558452}, the task assignment can be done in the task planning unit. 
\end{itemize}

	\begin{figure}[]
	\begin{center}
		\includegraphics[width=7.5cm]{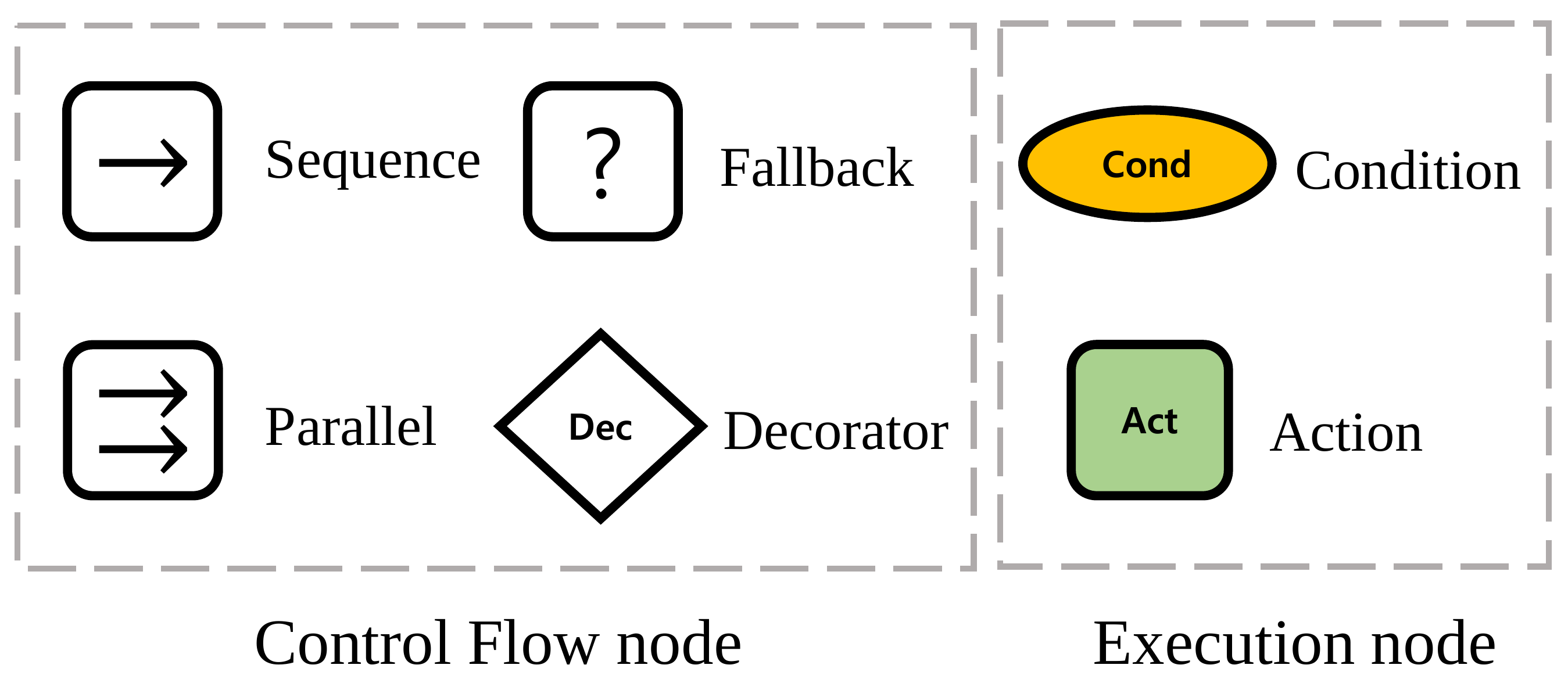}
		\caption{\label{standardbt}Standard BT nodes}
		\vspace{-0.5cm}
	\end{center}
\end{figure}

\section{Methods}
\subsection{Brief Introduction of BTs}

A BT starts from a root node that transmits ticks with a certain frequency to child nodes. When a child node receives a tick from the parent node, the result of the child node is fed back to the parent node based on its return status; success, failure, or running. The parent node then determines which child nodes are to be executed according to the results. Child nodes are largely divided into two types: control flow nodes and execution nodes.
Control flow nodes are further divided into four specific types, namely sequence, fallback, parallel, and decorator nodes. 
For a specific task moving through the control flow, another type of node is presented, namely an execution node. There are two types of execution nodes, namely condition and action nodes. Execution nodes differ from control flow nodes in that they are located at the end of the BT and have no child nodes. Therefore, they are also called leaf nodes.

\textbf{Root}: One BT has one root node. A root node ticks to its child nodes and all child nodes in a BT share the same clock frequency from their root node.

\textbf{Sequence}: A sequence node has one or more child nodes. If a child node returns a success, a sequence node ticks to the remaining child nodes. The graphical symbol for this node is a right-pointing arrow.

\textbf{Fallback}: In contrast to a sequence node, a fallback node sends a tick to the next child node when a child node returns a failure. The graphical symbol for this node is a question mark.

\textbf{Parallel}: Unlike sequence or fallback nodes, which only execute one child node sequentially, a parallel node executes multiple child nodes simultaneously. It can execute child nodes infinitely or a given number of times. If more than $M$ child nodes fail, then this node returns a failure. A parallel node is suitable for the parallel processing of specific tasks. The graphical symbol for this node is two right-pointing arrows.

\textbf{Decorator}: This node is placed on top of other nodes to decorate their child nodes. The decorator node is used to assist with the configuration settings of child nodes. The graphical symbol for this node is a diamond.

\textbf{Action}: This node receives a tick from a parent node such as a control flow node and completes an action. The action node returns a failure if it does not succeed completely. Unlike other nodes, an action node can return to the running state in addition to returning a success or failure. Since the execution process continues until the action is completed, the action state is considered to be running. The graphical symbol for this node is the associated action name.

\textbf{Condition}: As a node for expressing \textit{if-then} statements similar to traditional programming languages, the condition node is very helpful when the robot needs to be reactive.

The control flow nodes and execution nodes should be executed on only one specific type of robot hardware. This is why the standard BT in Fig. \ref{standardbt} is sufficiently strong to control a single robot. Therefore, special treatment of the standard BT is required to handle multiple robots. To distribute standard BT nodes to several robots in a simple manner, they should be able to send and receive ticks from parent or child nodes located on other robots. Therefore, an excessive amount of bandwidth must be exhausted for transmitting ticks and results through communication networks. Without loss of generality, multiple mobile robots are more likely to communicate through wireless networks, where communication uncertainty frequently occurs. Therefore, a novel BT node type that uses less bandwidth and has less uncertainty is required for multiple mobile robots.
\subsection{Data distribution service}

\begin{figure}[]
	\begin{center}
		\includegraphics[width=8.5cm]{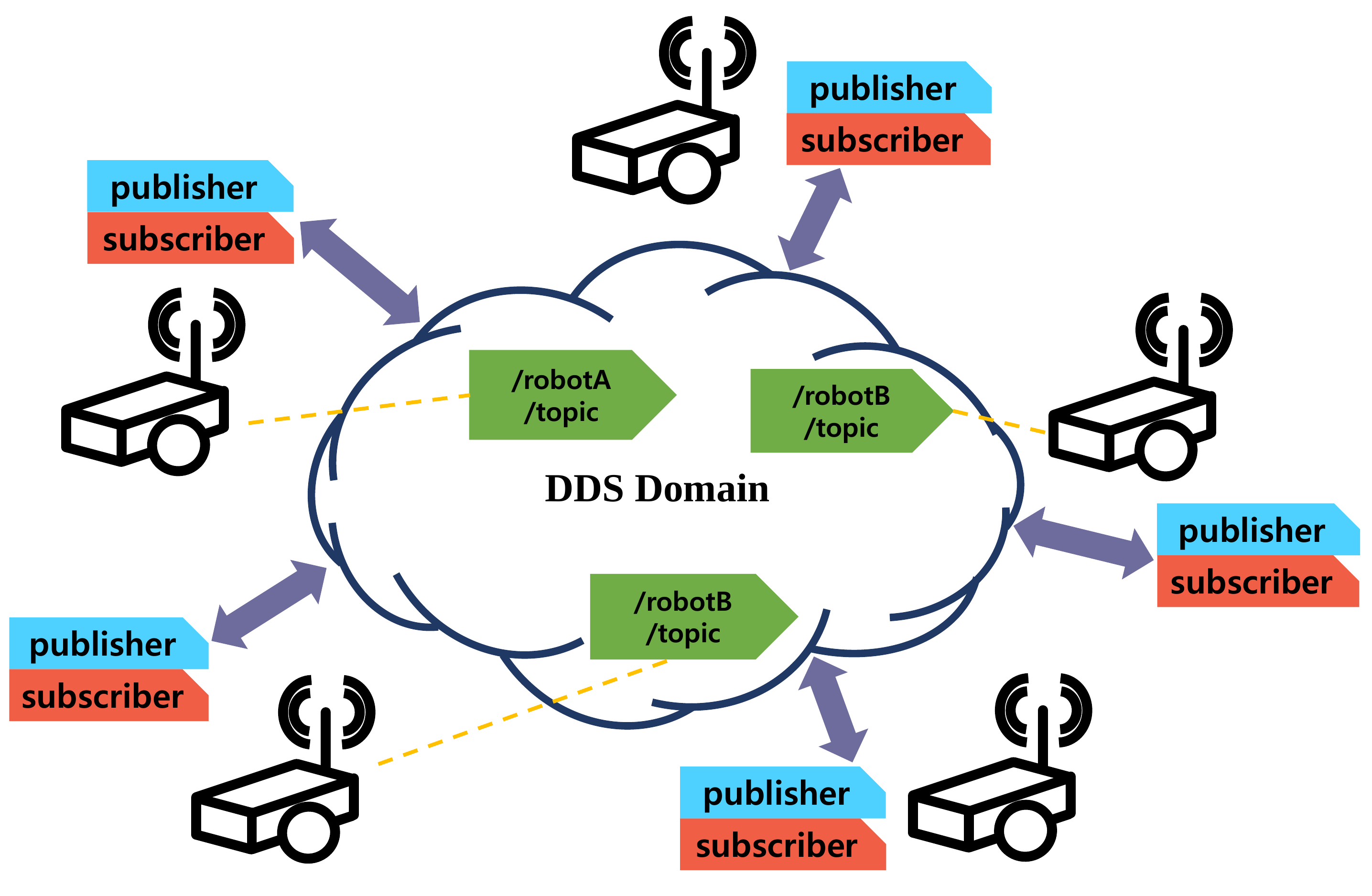}
		\caption{Data sharing in the global data space in DDS}
		\label{ddsconcept}
		\vspace{-0.5cm}
	\end{center}
\end{figure}

In distributed system environments, a DDS is data-centric and all participating robots in the DDS domain are allowed to be both data publishers and data subscribers as depicted in Fig. \ref{ddsconcept}. By writing data to the same DDS domain from any robot, additional participating robots can read the data. In a DDS, the standard data type is described by a topic name and value pair $topic=<name, value>$. All data are virtually stored in a global data space. The global data space may be in a distributed set of hardware units or the local memory of individual robots. Rather than determining whether data are stored on hardware units or a specific robot, the data is reliably shared with multi-cast way. To identify which topics are involved, an identifier such as a namespace is prefixed to the topic name in the following style: $/namespace/topic$. A namespace typically corresponds to the name of a robot. 

In previous studies \cite{4097991, 891049, 1372526}, a single centralized server would send commands to robots through a communication network to control multiple robots. Each robot subsequently determined the information of other robots through a single centralized server. If all robots operated without any problems, then control through a server provides adequate performance. However, if a robot's battery runs out, then the powered-down robot cannot share its position and may collide with other robots. In severe cases, the system may fall into a deadlock state if several robots lose their connection to the server. To prevent this issue, a DDS adopts dynamic discovery based on multicasting. With dynamic discovery, a robot application no longer requires the definition of communication endpoints because all robots can enter or exit the DDS domain dynamically. DDS middleware always attempts to discover the robots participating in the domain, rather than searching for other robots using a specialized robot application in advance.

\subsection{Actions in a DDS}
Since a DDS is data-centric, an action in a BT should be defined in the form of data on the corresponding DDS, where an action is represented in the form of variables, rather than functions.

\textbf{Definition 1} (Action in a DDS): An action $\mathcal{A} \in \mathcal{T}_{l}$ is a four tuple
$
\mathcal{A}=\left\{c, x, r, \Delta t\right\},
$
where $l \in \mathbb{N}$ is the index of the tree, $c$ is the command, $x$ is the state, and $r: \mathbb{R}^{n} \rightarrow\{\mathcal{R}, \mathcal{S}, \mathcal{F}\}$ is the return status; Running $(\mathcal{R})$, success $(\mathcal{S})$ and failure $(\mathcal{F})$. $\Delta t$ denotes the time step generated by the tick. 

\textbf{Definition 2} (Action variable): An action variable ${a}_{i}$ is a three tuple
$ 
{a}_{i}=\left\{c_{i}, x_{i}, r_{i}\right\},
$ 
where $i \in \mathbb{N}$ is the index of action $\mathcal{A}$, $c_{i}$ is the $i$-th known command, $x_{i}$ is the $i$-th unknown state, and $r_{i}$ is the $i$-th unknown return status.

\textbf{Definition 3} (Dual action variable): For any action variable ${a}_{i}$, another dual ($d$) action variable ${a}_{i}^d$ is presented.
$
{a}_{i}^d=\left\{c_{i}^d, x_{i}^d, r_{i}^d\right\},$ 
where $c_{i}^d$ denotes the $i$-th unknown dual command, $x_{i}^d$ denotes the $i$-th known dual state, and $r_{i}^d$ denotes the $i$-th known dual return status.

\textbf{Definition 4} (Intermediary memory variable): For an action variable and dual action variable, assume that there is also an intermediary memory ($m$) variable $a_{i}^{m}$;
\begin{equation*}
{a}_{i}^m=\left\{c_{i}^m, x_{i}^m, r_{i}^m\right\}.
\end{equation*}
All tuple shapes are the same with $a_{i}$ and $a_{i}^d$, and the tuples are all known.
By storing each known tuple in an intermediary memory variable and propagating the tuples of an intermediary memory variable to the extra unknown tuple of both $a_{i}$ and $a_{i}^d$, $a_{i}$ and $a_{i}^d$ become mutually known as follows:
	
\begin{equation*}
\begin{aligned}
&c_{i}^d \gets c_{i}^m \gets c_{i},\\
&x_{i} \gets x_{i}^m \gets x_{i}^d, \\
&r_{i} \gets r_{i}^m \gets r_{i}^d.
\end{aligned}
\end{equation*}

\textbf{Definition 5} (Dual Action): For a dual-action variable, a dual action in a BT can be created.
\begin{equation*}
\mathcal{A}^d=\left\{c^d, x^d, r^d, \Delta t\right\}
\end{equation*}
By assigning a dual action to another BT, the dual action with a non-equivalent tick $\Delta t'$ can be defined as follows:
\begin{equation*}
\mathcal{A}^d=\left\{c^d, x^d, r^d, \Delta t'\right\}.
\end{equation*}

\subsubsection {Action client \& server}
We now define an agent with an action on the action client and another agent with a dual action on the action server through parallelism. The action client and server each have local physical memory space for storing a variable that is transmitted to the intermediary memory variable from non-uniform ticks. Algorithms \ref{alg:actionclient} and \ref{alg:actionserver} define the processes involved in executing an action.

\begin{algorithm}[t]
	\caption{Action Client}
	\begin{algorithmic}
		\Function{Tick}{\null}
			\State Publish command $c_{i}$ to $c_{i}^m$
			\State Request command process from the server
			\State Wait for the response of the action server
		\If {response == accepted}
			\State {$x_{i} \gets x_{i}^m$} \Comment{Subscribe $x_{i}^m$ to $x_{i}$}
			\State  {$r_{i} \gets r_{i}^m$} \Comment{Subscribe $r_{i}^m$ to $r_{i}$}
		\If {$r_{i}==\mathcal{S}$} \State \Return {$\mathcal{S}$}
		\ElsIf {$r_{i}==\mathcal{F}$} \State \Return {$\mathcal{F}$}
		\Else \State \Return {$\mathcal{R}$}
		\EndIf	
		\Else
			\State \Return {$\mathcal{F}$}
		\EndIf
		\EndFunction
	\end{algorithmic}
	\label{alg:actionclient}
\end{algorithm}

The actions in a BT can be executed by a single robot. To apply an action to multiple robots, the action is split into a set of action client and server. The action client sends a command to the DDS domain and the action server receives and executes the command on a real robot. The DDS action consists of three parts: a command that specifies the target task, a state that indicates the current state of the robot, and a return status that indicates whether the action has been completed, where the result returns $\mathcal{S}$, $\mathcal{F}$, and $\mathcal{R}$, similar to the traditional BT action.

\subsubsection {Delay-free accessibility} An intermediary variable is delay-free accessible with respect to both the action client and server. The action client and server do not read or write the variable simultaneously. One side writes the variable and the other side reads the variable. Delay-free access to an intermediary variable contributes to real-time communication.

\begin{algorithm}[t]
	\caption{Action Server}
	\begin{algorithmic}
		\Function{Tick}{\null}
		\State Get the action client request  
		\If {request is acceptable}
		\State {response $\gets$ accepted}
		\State {Send the response to the action client}
		\State {$c_{i}^d \gets c_{i}^m$} 
		\If {command != complete}
		\State {$x_{i}^d = ExecuteCommand()$}
		\State {$x_{i}^m \gets x_{i}^d$} \Comment{Publish $x_{i}^d$ to $x_{i}^m$}
		\State {$r_{i}^d \gets \mathcal{R}$}
		\Else
		\State {$r_{i}^{d} \gets \mathcal{S}$}
		\EndIf
		\Else
		\State {response $\gets$ command-rejected}
		\State {Send the response to the action client}
		\State  {$r_{i}^d \gets \mathcal{F}$}
		\EndIf
		\State {$r_{i}^m \gets r_{i}^d$} \Comment{Publish $r_{i}^d$ to $r_{i}^m$}
		\EndFunction
	\end{algorithmic}
	\label{alg:actionserver}
\end{algorithm}

\subsection{BT nodes for connecting split action}
\begin{figure}[]
	\begin{center}
		\includegraphics[width=6.5cm]{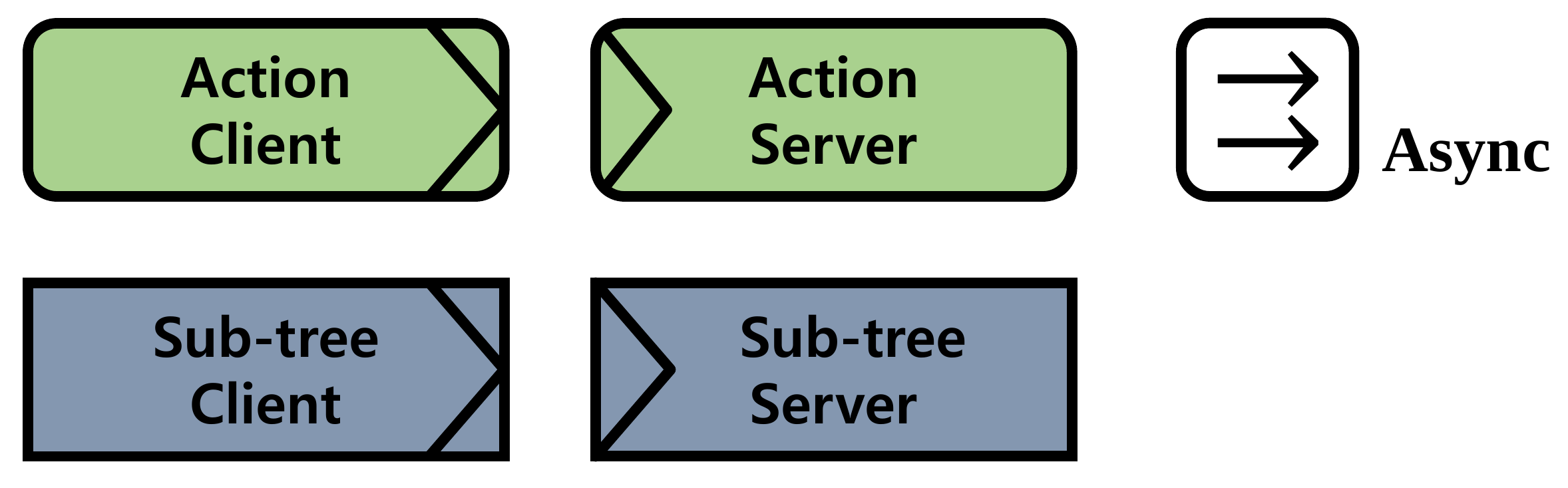}
		\caption{Action client, server, asynchronous parallel node, sub-tree client, and sub-tree server}
		\label{newtype}
		\vspace{-0.5cm}
	\end{center}
\end{figure}

\textbf{Definition 5} (Asynchronous parallel node): For more than one BT, if the BTs do not share a tick and thus do not communicate with intermediary variables, then the BTs are considered to be asynchronously parallel. An asynchronous parallel node has at least one action client and one action server. If a sub-tree of a BT has the action client or server, then a sub-tree becomes the sub-tree client or server. The symbol for this node is shown in Fig. \ref{newtype}.

\textbf{Definition 6} (Asynchronous parallel BT): Each BT has at least one action node. A BT is allowed to split into two or more sets of subtree clients and servers, as indicated by the two rectangular symbols in Fig \ref{newtype}. Then, the BT gains more than two root nodes. By adding nonequivalent ticks to BTs, each BT can be the child node of an asynchronous parallel BT.

Any action that has an action variable can produce a dual action. Each type of action communicates with the intermediary variable $a_{i}^d$. However, the variable is delay free and accessible. This indicates that the variable is not mutually exclusive, but shareable with respect to asynchronous parallel BTs.

\subsubsection {Sequence \& fallback node split} 
Let a BT have ${\mathcal{T}_{l}}$ and ${\mathcal{T}_{m}}$ sub-trees. $\mathcal{T}_{l}$ can be then split into ${\mathcal{T}_{l}}_{client}$ and ${\mathcal{T}_{l}}_{server}$. In addition, $\mathcal{T}_{sequence}$ or $\mathcal{T}_{fallback}$ is converted to the asynchronous parallel BT which has the sequence or fallback node in the following manner.

{\small
\begin{equation}
\begin{aligned}
\mathcal{T}_{sequence} &= Sequence(\mathcal{T}_{l}, \mathcal{T}_{m}) \\
		&= AsyncParallel({\mathcal{T}_{l}}_{client}, Sequence({\mathcal{T}_{l}}_{server}, \mathcal{T}_{m})) \\
\mathcal{T}_{fallback} &= Fallback(\mathcal{T}_{l}, \mathcal{T}_{m}) \\
&= AsyncParallel({\mathcal{T}_{l}}_{client}, Fallback({\mathcal{T}_{l}}_{server}, \mathcal{T}_{m}))		
\end{aligned}
\end{equation} }
\subsubsection{Multiple sequence \& fallback node split} 
A bundle of sequence or fallback nodes can be split into an asynchronous parallel BT.

{\small
\begin{equation}
\begin{aligned}
&\prod_{i=1}^{N} \mathcal{T}_{sequence}^{i} :=\left\{\mathcal{T}_{sequence}^{1}, \mathcal{T}_{sequence}^{2}, \cdots , \mathcal{T}_{sequence}^{N}\right\} \\
&=\prod_{i=1}^{N}  AsyncParallel({\mathcal{T}_{l}^{i}}_{client}, Sequence({\mathcal{T}_{l}^{i}}_{server}, \mathcal{T}_{m}^{i})) \\
&=AsyncParallel(\prod_{i=1}^{N} {\mathcal{T}_{l}^{i}}_{client}, \prod_{i=1}^{N} Sequence({\mathcal{T}_{l}^{i}}_{server}, \mathcal{T}_{m}^{i})) \\
&\prod_{i=1}^{N} \mathcal{T}_{fallback}^{i} =\left\{\mathcal{T}_{fallback}^{1}, \mathcal{T}_{fallback}^{2}, ..., \mathcal{T}_{fallback}^{N}\right\} \\
&=\prod_{i=1}^{N}  AsyncParallel({\mathcal{T}_{l}^{i}}_{client}, Fallback({\mathcal{T}_{l}^{i}}_{server}, \mathcal{T}_{m}^{i})) \\
&=AsyncParallel(\prod_{i=1}^{N} {\mathcal{T}_{l}^{i}}_{client}, \prod_{i=1}^{N} Fallback({\mathcal{T}_{l}^{i}}_{server}, \mathcal{T}_{m}^{i}))
\end{aligned}
\end{equation}
}
\vspace{-0.5cm}
\subsection{Applying new types of BTs to robots}
\begin{figure}[]
	\begin{center}
		\includegraphics[width=4.5cm]{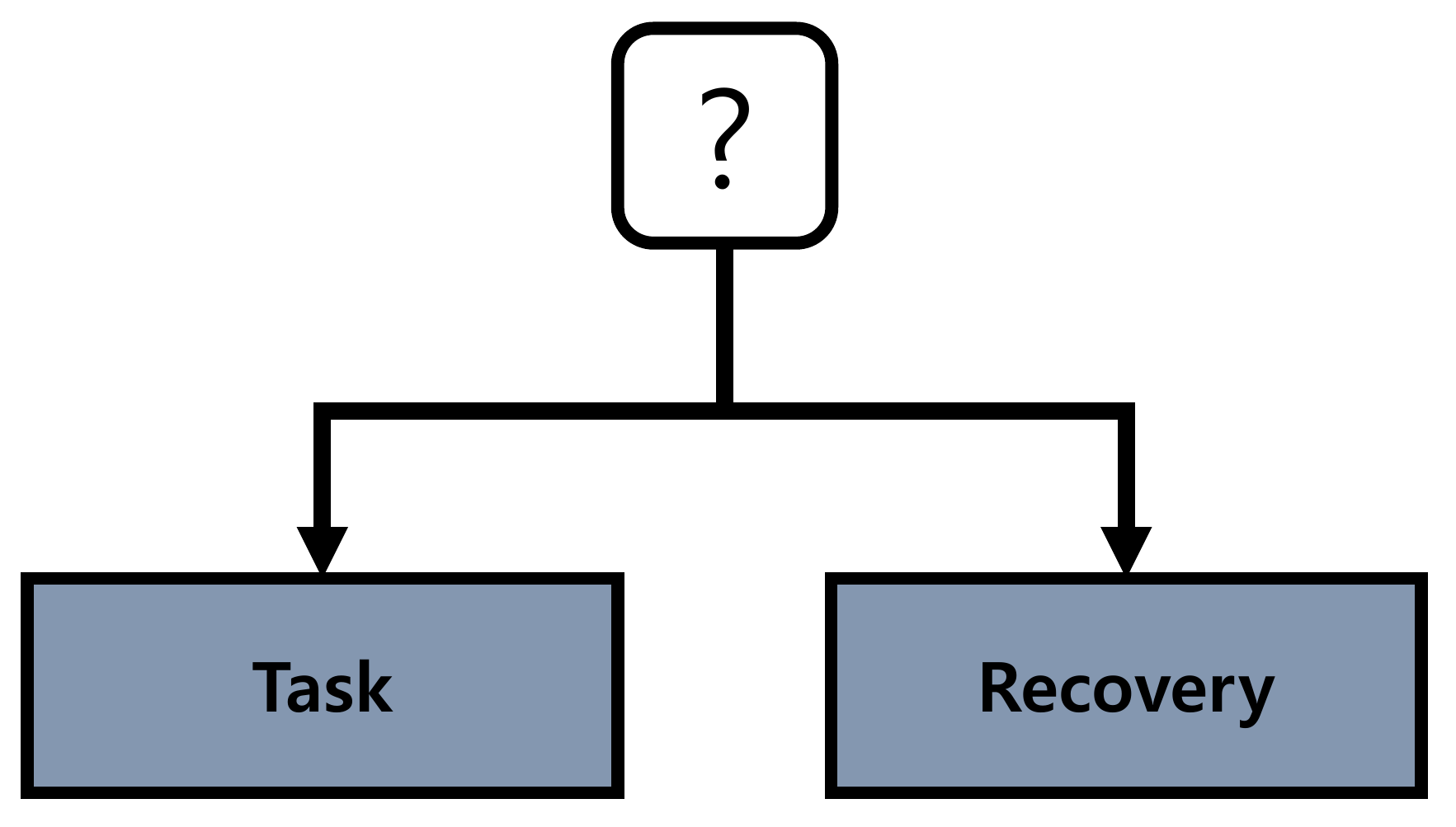}
		\caption{Typical BT topology for a single robot}
		\label{singlerobotbt}
		\vspace{-0.5cm}
	\end{center}
\end{figure}

\begin{figure}[thb]
	\begin{center}
		\includegraphics[width=7.5cm]{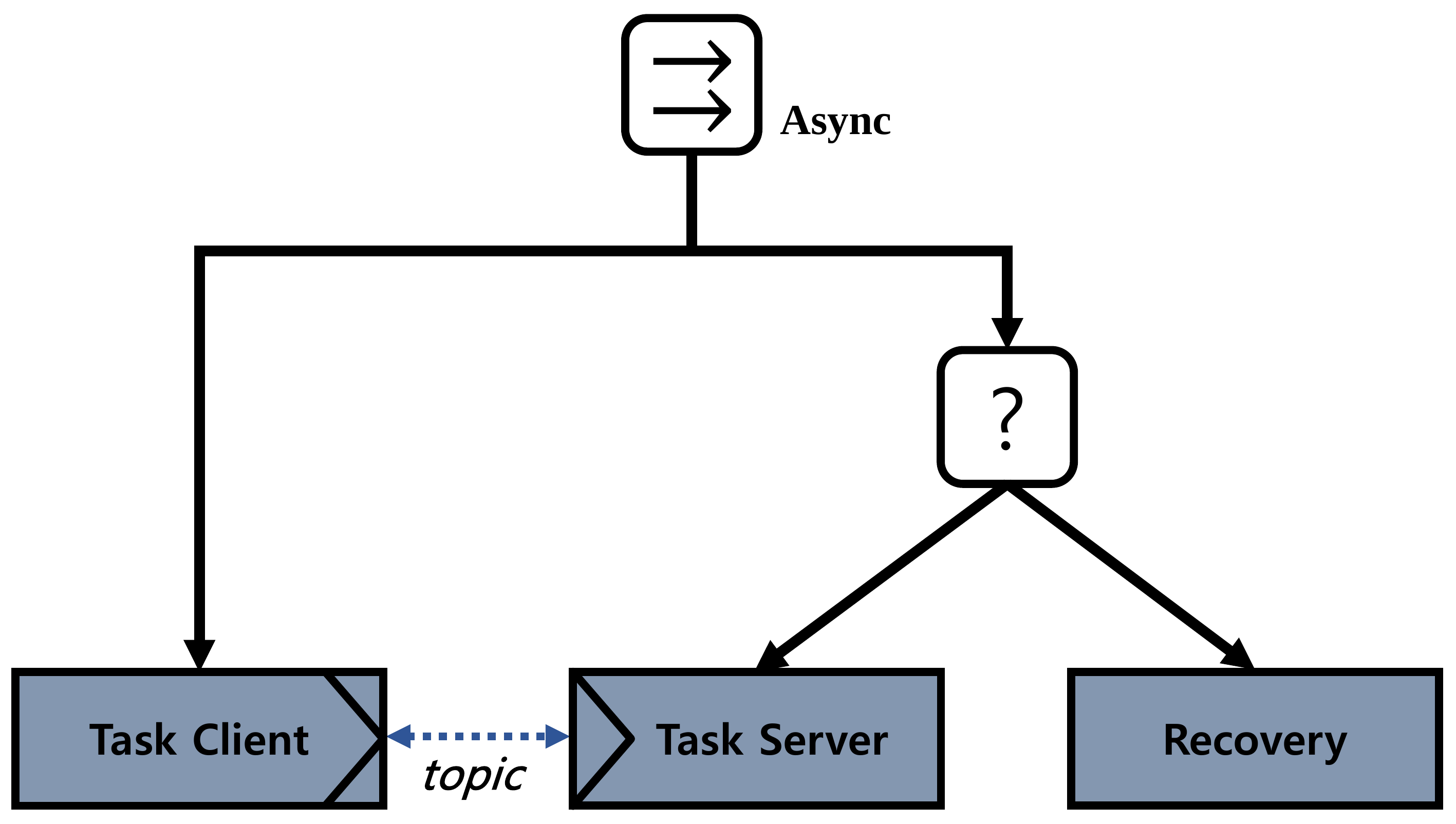}
		\caption{A task action split into a task action client \& server}
		\label{clientserverbt}
	\end{center}
\end{figure}

A BT topology suitable for a single robot is presented in Fig. \ref{singlerobotbt}.
When a robot is working with a given task BT ${\mathcal{T}_{task}}$, it typically enters the recovery mode when it encounters an undesirable and undetermined scenario. Therefore, the recovery BT ${\mathcal{T}_{recovery}}$ for the fault-tolerant function of ${\mathcal{T}_{task}}$ is tied to the fallback node. 
However, if the task is not known to the robot in advance, it is necessary to learn the task to execute the BT. Let the unknown task node take charge of the task server ${\mathcal{T}_{task}}_{server}$ and build up the known task of $\mathcal{T}_{task}$ as a task client ${\mathcal{T}_{task}}_{client}$ before sending the task.
In other words, ${\mathcal{T}_{task}}$ is split into ${\mathcal{T}_{task}}_{client}$ and ${\mathcal{T}_{task}}_{server}$, as shown in Fig. \ref{clientserverbt}.
By assigning ${\mathcal{T}_{task}}_{client}$ to a task planner on separated hardware, ${\mathcal{T}_{task}}_{server}$ is capable of sending the state and results in ${\mathcal{T}_{task}}_{client}$ using the following formula:
\begin{equation}
\begin{aligned}
\mathcal{T}_{fallback} &= Fallback(\mathcal{T}_{task}^{j}, \mathcal{T}_{recovery}^{j}) \\
&= AsyncParallel({\mathcal{T}_{task}^{i}}_{client}, \\ &Fallback({\mathcal{T}_{task}^{j}}_{server}, {\mathcal{T}_{recovery}^{j}})),		
\end{aligned}
\label{eqn:totalasync}
\end{equation} 
where $i$ is a task planner and $j$ is a single robot.

Both ${\mathcal{T}_{task}}_{client}$ and ${\mathcal{T}_{task}}_{server}$ have the action.
Then, the action tuples must be transmitted to a BT in which the action is not clearly defined and must be changed variably. To allow several BTs to communicate with each other, action tuples between more than one BT should be transmitted to the global space according to the DDS topic. The DDS topic is shared by individual robots and the task planner in the DDS domain.

\section{Experimental results}
\subsection{Hardware and  software setup}
\begin{figure}[]
	\begin{center}
		\includegraphics[width=6.5cm]{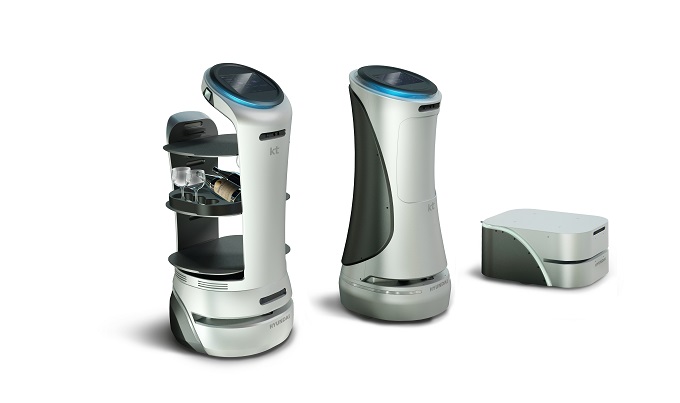}
		\caption{Hyundai Robotics mobile robot lineup. The left two robots have 50 kg payloads and the right robot has a 100 kg payload.}
		\label{hyundairobots}
		\vspace{-0.5cm}
	\end{center}
\end{figure}

\begin{figure}[h]
	\begin{center}
		\includegraphics[width=7cm]{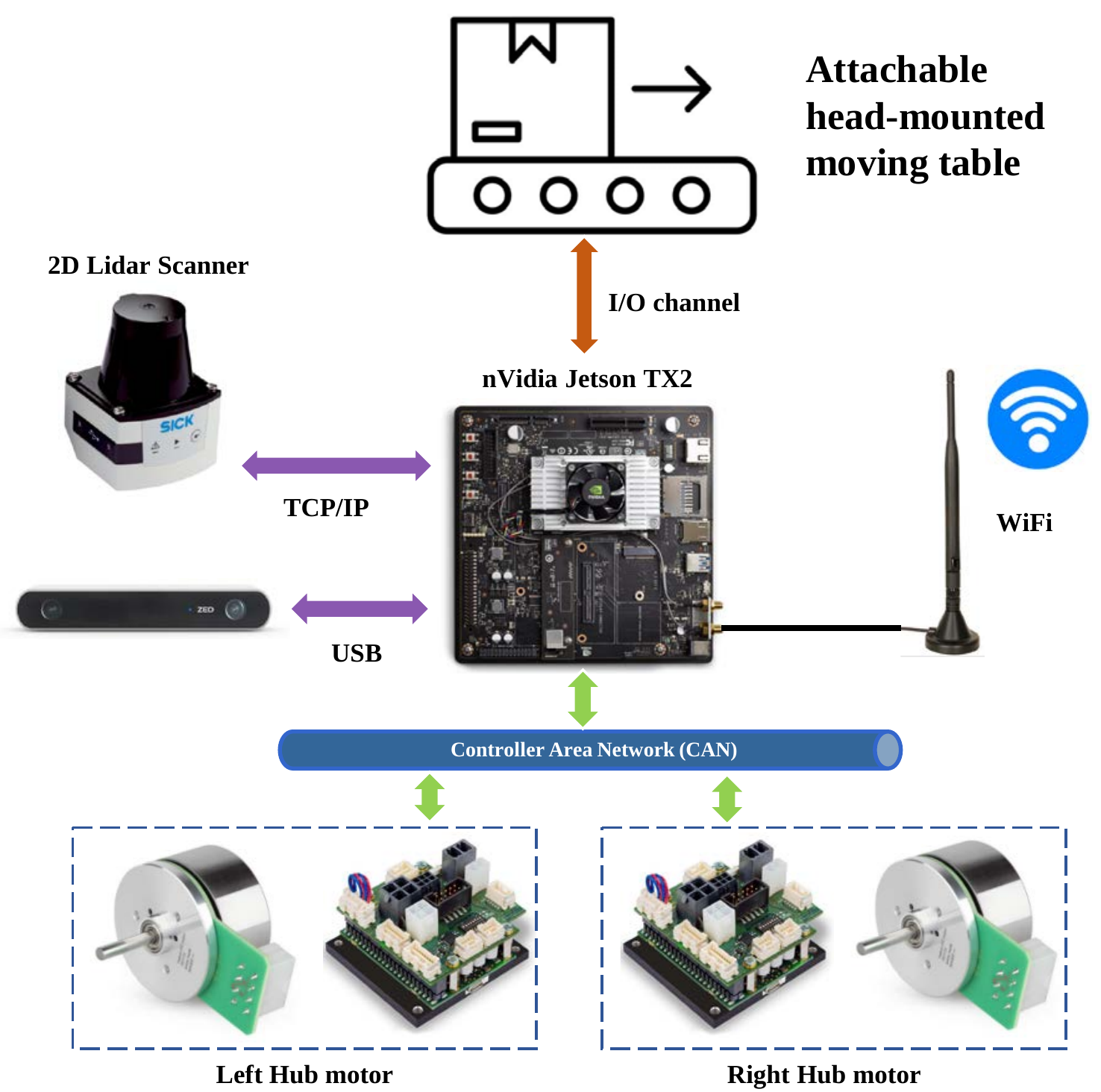}	
		\caption{Topology of devices attached to the mobile robot computer}
		\label{hardwaretopology}
	\end{center}
\end{figure}

The test robots in Fig. \ref{hyundairobots} used in this study were developed by Hyundai Robotics. The robots have a differential drive, Wi-Fi for the DDS, sensors for tracking equipment, and a peripheral input/output (I/O) module for delivery. The differential drive is an all-in-one motor containing a brushless motor, motor drive on a controller area network, wheels, and tires. The tracking sensors consist are a 2D lidar scanner and a stereo vision camera for positional sensing and obstacle detection. The peripheral I/O module can be connected to a rotatable or prismatic actuator. All devices above is controlled to the mobile robot computer. The topology of devices is depicted in Fig. \ref{hardwaretopology}. It is mainly aimed at transportation in factories or providing simple delivery services to customers in hotels.

The robot performs global and local path planning on an occupancy grid map. The occupancy grid map was downloaded on multiple robots before planning. Once the global path planning for a given goal is completed, the robot performs path tracking. The goal of the BT for the robots is to transmit information through a DDS over Wi-Fi. The DDS library was taken from FastDDS\footnote{https://github.com/eProsima/Fast-DDS} and the BT library was taken from BehaviorTree.CPP\footnote{https://www.behaviortree.dev/}. The action client and server library is the action in robot operating system 2 (ROS 2) Eloquent\footnote{https://docs.ros.org/en/eloquent/Tutorials/Understanding-ROS2-Actions.html}. 

\begin{figure}[]
	\begin{center}
		\includegraphics[width=8.5cm]{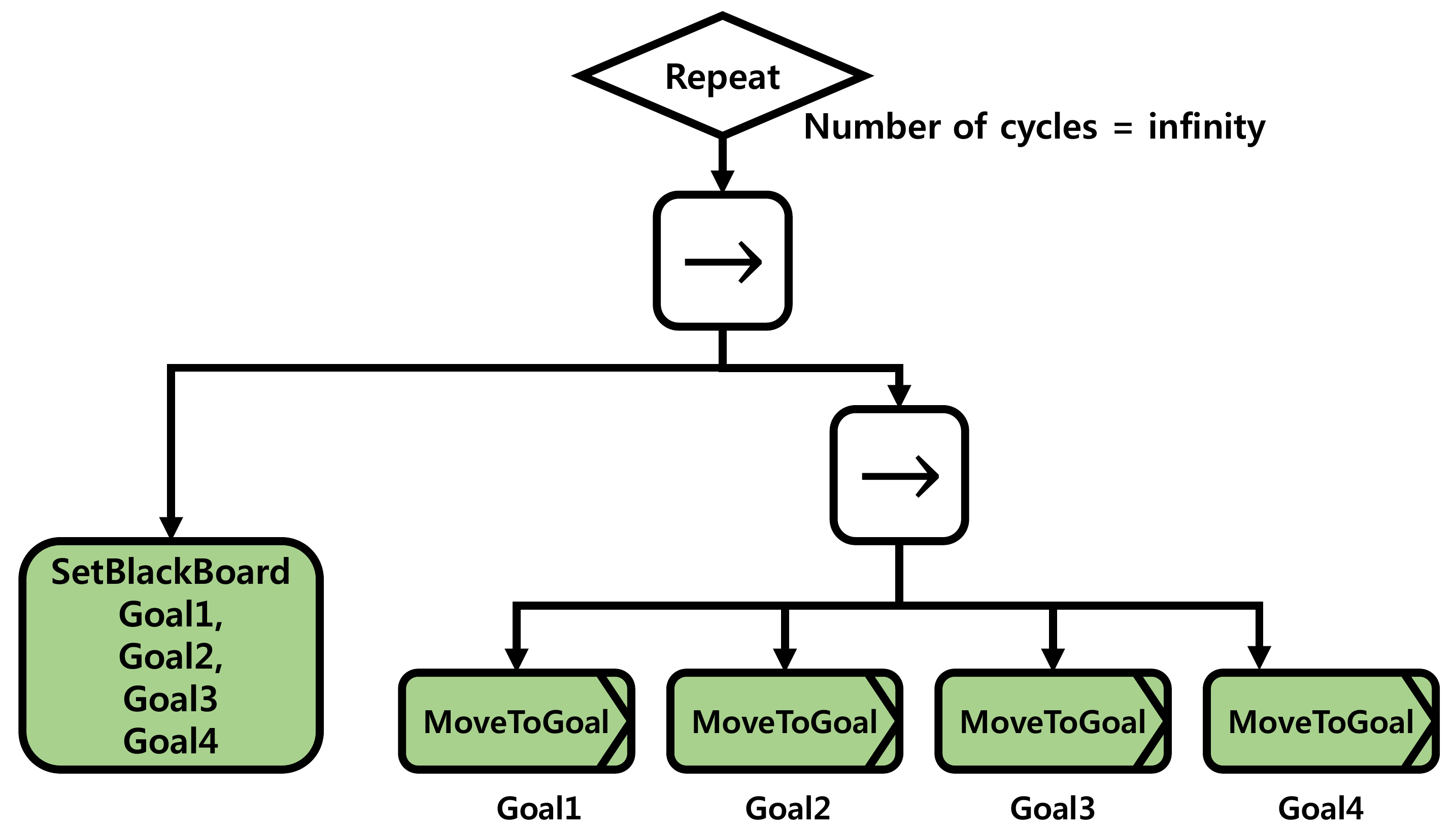}
		\caption{Task planning BT}
		\label{taskbt}
	\end{center}
\end{figure}
\begin{figure}[h]
	\begin{center}
		\includegraphics[width=7.5cm]{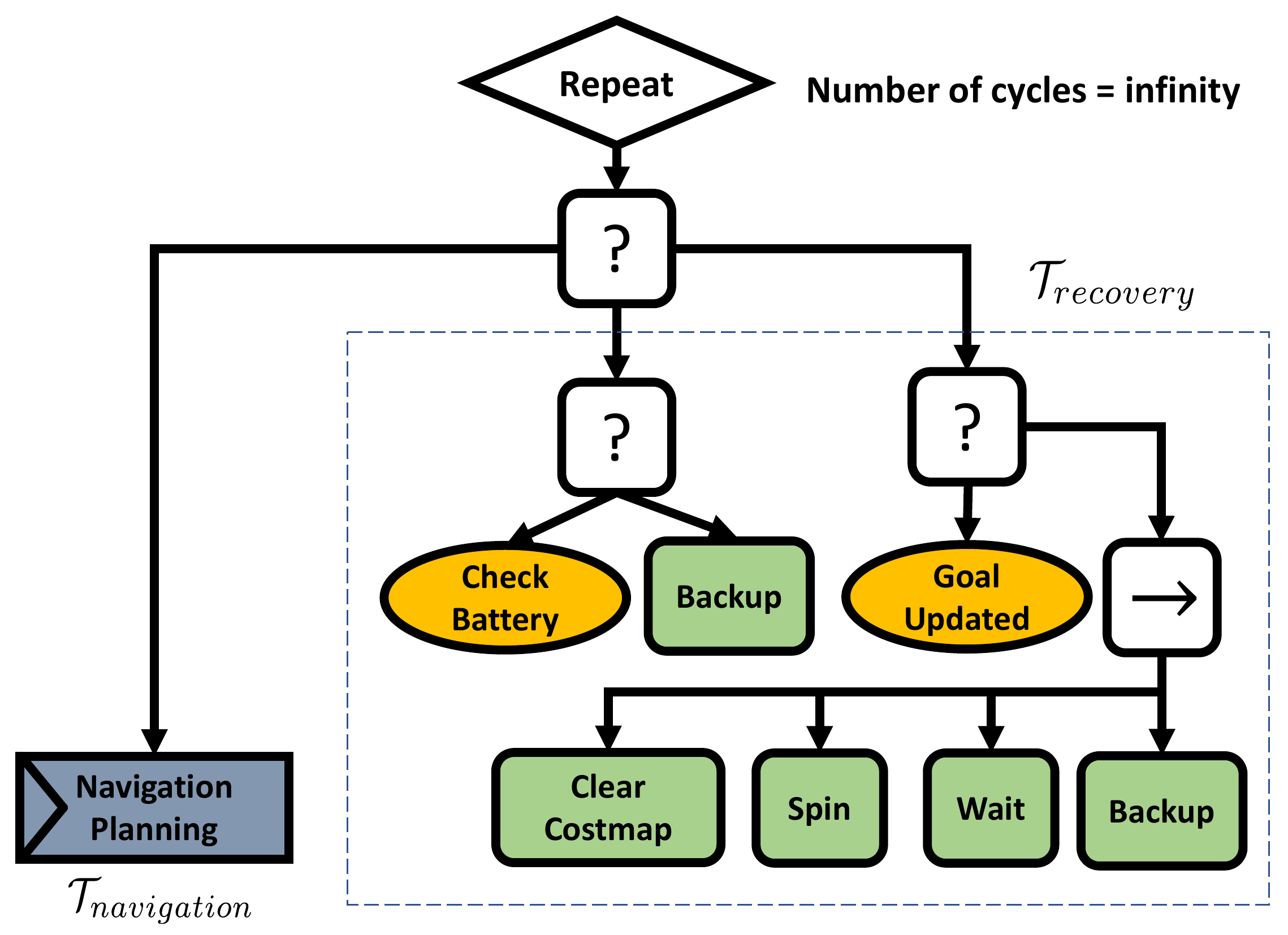}
		\caption{Navigation planning \& recovery BT in a robot}
		\label{recoverybt}
	\end{center}
\end{figure}
\begin{figure}[h]
	\begin{center}
		\includegraphics[width=7.5cm]{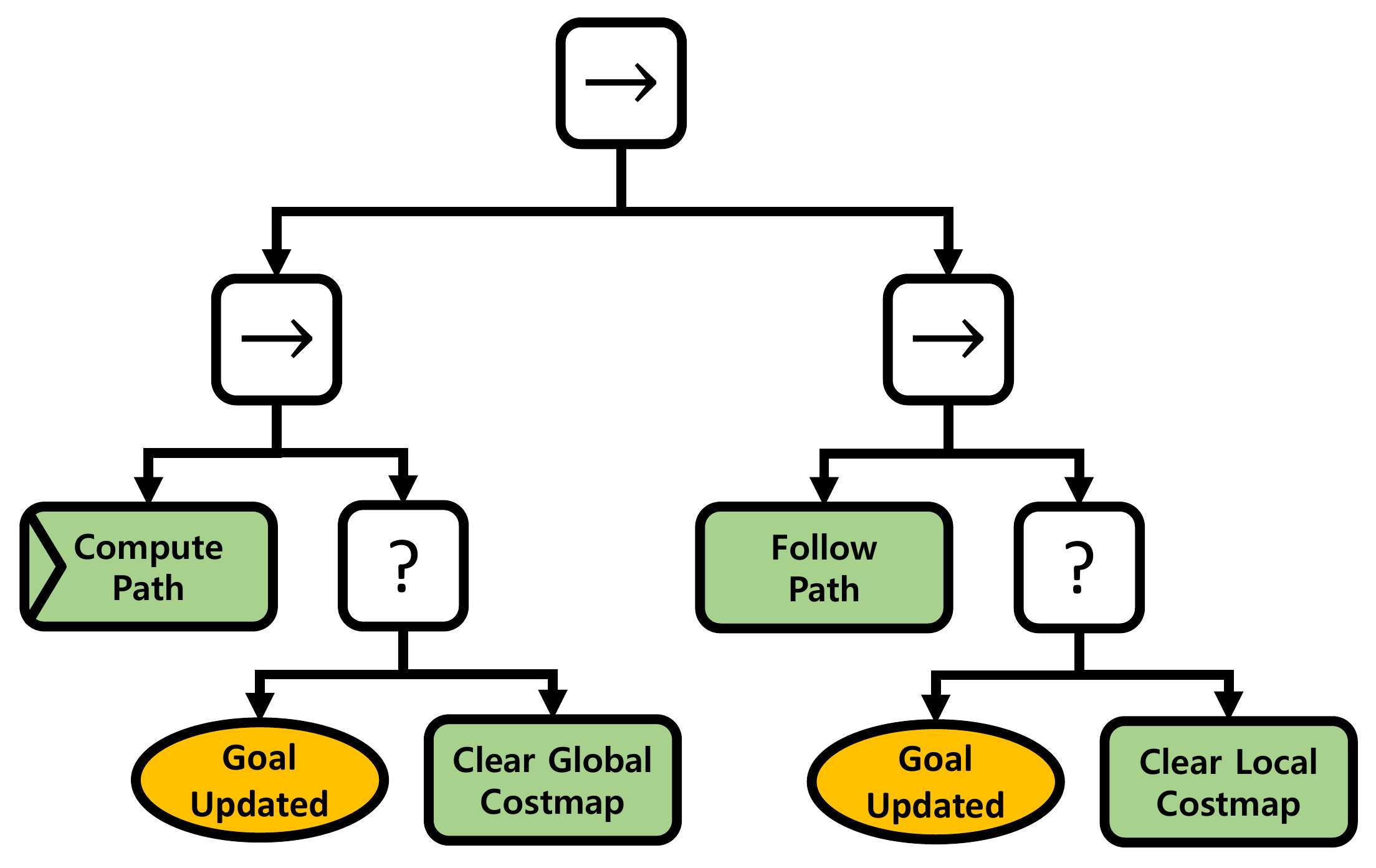}
		\caption{Navigation planning BT}
		\label{navigatewithplanning}
	\end{center}
\end{figure}

\subsection{Basic BT setup for a robot with a task planner}
The task planning BT is presented in Fig. \ref{taskbt}. It was assumed that the robot task in this study was the simple movement to the designated goal point. Therefore, the specific command to be sent to the action server was the goal point coordinate in the form of a topic. Whenever ticks appear in ${\mathcal{T}_{navigation}}$ in Fig.~\ref{navigatewithplanning}, the goal point is read from the action server in ${\mathcal{T}_{navigation}}$ as a topic and the robot performs global planning. If there are obstacles at the goal position or path planning fails as a result of obstacles in all directions, the recovery BT ${\mathcal{T}_{recovery}}$ under the fallback node should be executed, as shown in Fig. \ref{recoverybt}. All goal coordinates are stored in the task-planning BT with the \textit{SetBlackBoard} action in advance. Once the robot reaches the goal position without a global or local planning failure, the next goal is transmitted to ${\mathcal{T}_{navigation}}$. The task planning BT has four goal positions and the robot infinitely rotates on each of these goal position, as described by the \textit{Repeat} decorator node.

In ${\mathcal{T}_{navigation}}$, once global path planning is performed, the robots follow the planned paths. However, if planning fails, the global cost map is cleared. Similarly, if the robot cannot follow the path, the local cost map is cleared under the assumption of local planning failure. Before the global and local cost maps are cleared, we check whether the goal has been updated.

If ${\mathcal{T}_{navigation}}$ returns a failure, then ${\mathcal{T}_{recovery}}$ is finally executed. In the first stage, a battery check is performed. If the battery charge is low, then the robot returns to the backup position. If the battery condition is fair, then the robot finally clears the entire cost map. It then spins, waits for seconds.

\begin{figure*}[t]
	\begin{center}
		\includegraphics[width=18cm]{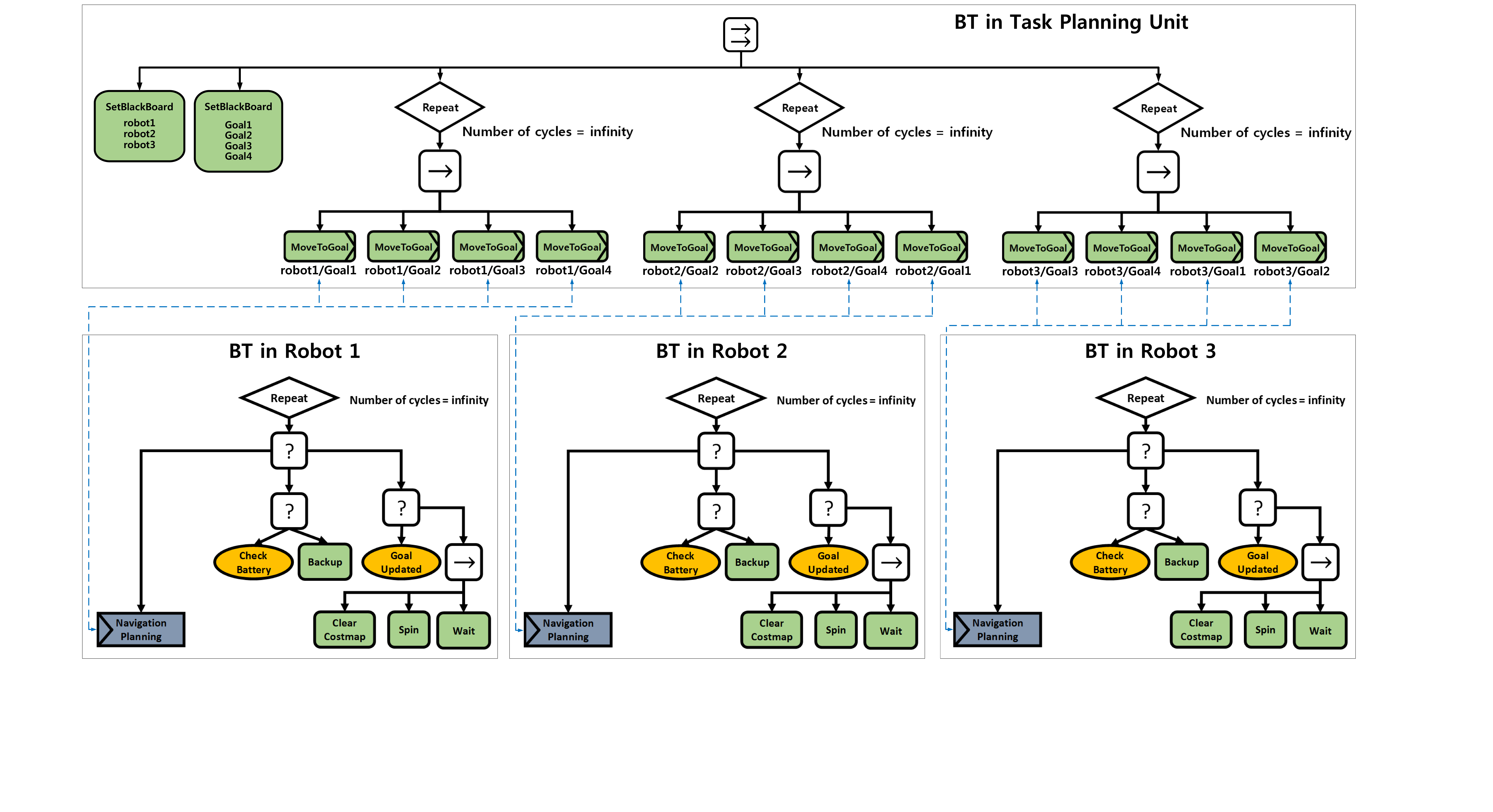}
		\caption{The BTs for the task planning unit and robots are separated. The BTs in the task planning unit are integrated into a single BT. Each action in the task planning unit has a topic. The topic for each action is written with the corresponding robot name plus the goal position. The topics are transmitted to the navigation planning subtree of the BT in each robot.}
		\label{multibt}
		\vspace{-0.5cm}
	\end{center}
\end{figure*}

\begin{figure}[btp]
	\begin{center}
		\includegraphics[width=8.5cm]{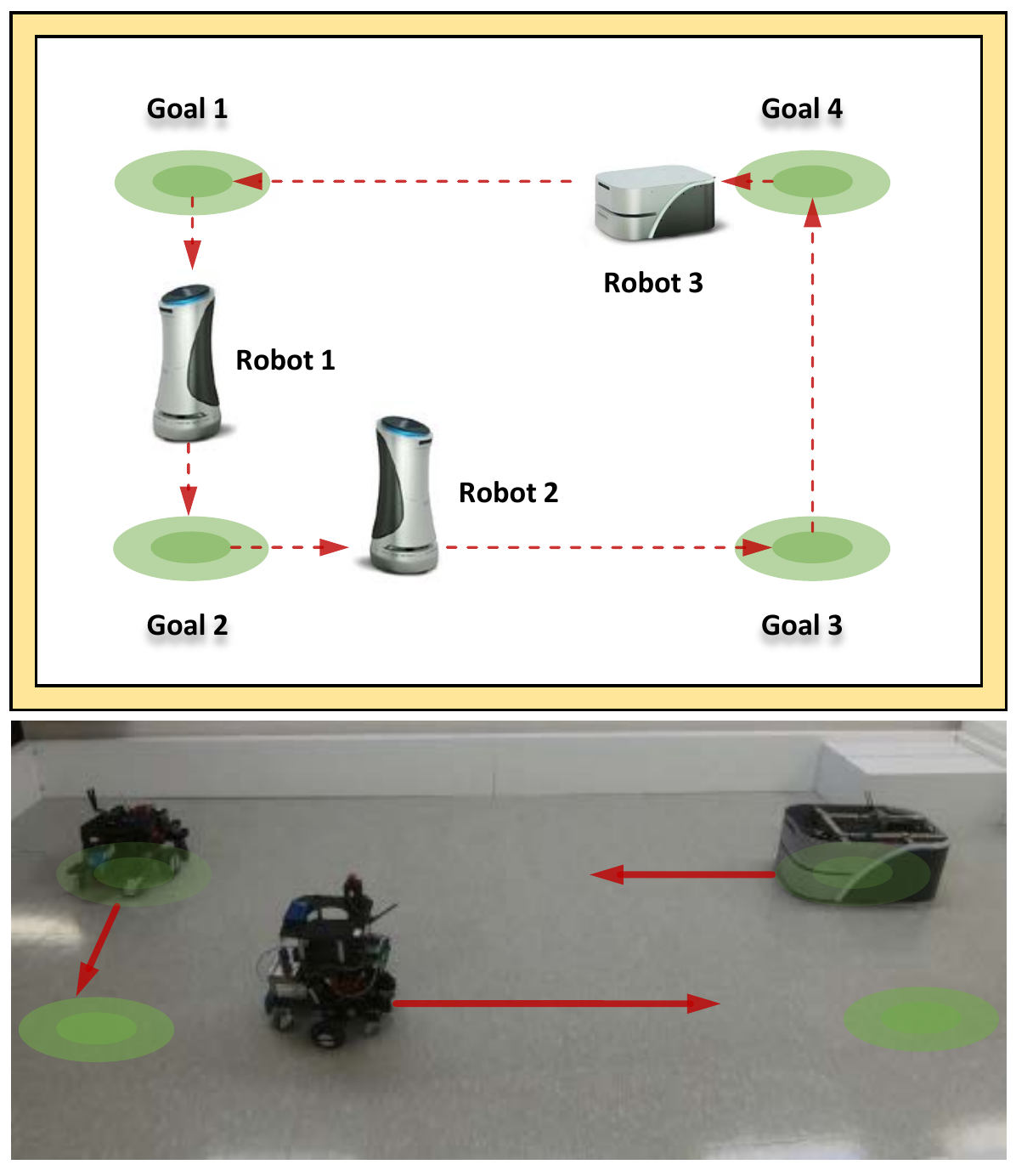}
		\caption {Three robots alternately traveling to four goal points. \newline Video link : https://www.youtube.com/watch?v=IFWfQtoWA34}
		\label{threerobots}
		\vspace{-0.5cm}
	\end{center}
\end{figure}

\subsection{Coalesced BT in the task planning unit for multiple mobile robots}
To control multiple robots, the robots and task planning unit must be clearly distinguished. The task planning unit delivers tasks to multiple robots and receives real-time feedback regarding whether the tasks are being performed through the communication network. 
To implement one task planning unit for multiple robots effectively, the task planning unit should be capable of dealing with multiple task BTs for individual robots at once. To execute task BTs in the task planning unit for $N$ robots, processes or threads for each task should be created. Each task in a process or thread has a single root node and different ticks are naturally generated. Rather than utilizing many ticks for independent BTs, one identical tick is more desirable for synchronizing multiple robots with one-rooted BTs, as shown in Fig. \ref{multibt}. A one-rooted BT connecting task planning BTs for multiple robots can be achieved using one parallel node according to the following equation:

\begin{equation}
\begin{aligned}
\prod_{i=1}^{N} {\mathcal{T}_{task}^{i}}_{client} = Parallel(\prod_{i=1}^{N} {\mathcal{T}_{task}^{i}}_{client}) = {\mathcal{T}_{multitask}^{tpu}}_{client},
\end{aligned}
\label{eqn:singleparallel}
\end{equation}
where $tpu$ denotes the task planning unit.
Each ${\mathcal{T}_{task}}_{client}$ is independent and $\prod_{i=1}^{N} {\mathcal{T}_{task}^{i}}_{client}$ can be coalesced into a single parallel BT node.
Then,  (\ref{eqn:totalasync}) becomes

\begin{equation}
\begin{aligned}
\prod_{i=1}^{N} \mathcal{T}_{fallback}^{i} = AsyncParallel({\mathcal{T}_{multitask}^{tpu}}_{client},\\
\prod_{j=1}^{N} Fallback({\mathcal{T}_{task}^{j}}_{server}, \mathcal{T}_{recovery}^{j})),
\end{aligned}
\end{equation}
where $i$ is the robot index and $j$ is the $j$-th robot.

To test whether the proposed BT framework works properly, we attempted to control three robots to move to four designated goal positions, as shown in Fig. \ref{threerobots}. The four positions marked by green circles are the vertices of a virtual rectangle. When one robot attempts to go to an adjacent vertex, the remaining robots also go to the designated vertex, as indicated by the red arrows. All robots repeat this process infinitely. We verified that the three robots were driven according to the actions and conditions set in the BTs presented in Fig. \ref{multibt}.

As shown in Fig. \ref{multibt}, all robots have a basic BT for navigation planning and their recovery mode. The BT in an individual robot shares its own variables with a single task planning unit. The single task planning unit only sends the goals to the intermediary variables with DDS topics in the global space. To distinguish DDS topics for individual robots, the topics are renamed in the $\slash robotname \slash topicname$ style. The BT in a single task planning unit starts from a parallel node derived from Equation (\ref{eqn:singleparallel}). The child nodes are independent of each other, but they can share blackboard variables. The blackboard variables included the four goal positions with coordinates.

\section{Conclusions}
In this paper, we proposed a BT-based asynchronous task planning for multiple mobile robots using a DDS.
Exploiting  the readability and flexibility of modularized BTs, we show that BTs for the task planning of multiple mobile robots can be integrated into a single BT in the task planning unit. The different purposes of BTs can be executed separately according to the BT split between an action client and a server.   This is the first step toward asynchronous task planning for multiple mobile robots using BTs.

\bibliographystyle{UNSRT}
\bibliography{references}

\end{document}